%% file: harmsmap.tex
\documentclass[manuscript]{acmart}
\pdfoutput=1

\usepackage{tabularx}
\usepackage{enumitem}
\usepackage{longtable}
\usepackage{xcolor}
\usepackage{wrapfig}
\usepackage{framed}
\usepackage{adjustbox}
\usepackage{tikz}
\usepackage{lipsum}
\usepackage{calc}
\newlength{\strutheight}
\usepackage{cleveref}
\crefname{section}{\S}{\S}
\crefname{Section}{\S}{\S}
\crefformat{section}{§#2#1#3}
\crefname{table}{Tab.}{Tab.}
\crefname{appendix_table}{Tab.}{Tab.}
\crefname{Table}{Tab.}{Tab.}
\crefname{Figure}{Fig.}{Fig.}
\crefname{figure}{Fig.}{Fig.}
\crefname{appendix}{Appendix}{Appendix}
\crefname{chapter}{Chapter}{Chapter}

\usepackage{subcaption}

\AtBeginDocument{%
  \providecommand\BibTeX{{%
    \normalfont B\kern-0.5em{\scshape i\kern-0.25em b}\kern-0.8em\TeX}}}

\setcopyright{rightsretained}
\acmYear{2023}
\acmDOI{XXXXXXX.XXXXXXX}

\acmConference[Assessing Language Model Deployment with Risk Cards]{}{April}{2023}

\acmISBN{978-1-4503-XXXX-X/18/06}

\begin{document}

\title{Assessing Language Model Deployment with Risk Cards}

\author{Leon Derczynski}
\email{leondz@uw.edu/ld@itu.dk}
\affiliation{%
  \institution{University of Washington/ITU Copenhagen}
  \country{USA/Denmark}
}
\author{Hannah Rose Kirk}
\email{hannah.kirk@oii.ox.ac.uk	}
\affiliation{%
  \institution{University of Oxford}
  \country{United Kingdom}
}
\author{Vidhisha Balachandran}
\email{vbalacha@cs.cmu.edu}
\affiliation{%
  \institution{Carnegie Mellon University	}
  \country{United States}
}
\author{Sachin Kumar}
\email{sachink@cs.cmu.edu}
\affiliation{%
  \institution{Carnegie Mellon University}
  \country{United States}
}
\author{Yulia Tsvetkov}
\email{yuliats@cs.washington.edu}
\affiliation{%
  \institution{University of Washington}
  \country{United States}
}
\author{M.R. Leiser}
\email{m.r.leiser@vu.nl}
\affiliation{%
  \institution{Vrije Universiteit Amsterdam}
  \country{Netherlands}
}
\author{Saif Mohammad}
\email{saif.mohammad@nrc-cnrc.gc.ca}
\affiliation{%
  \institution{National Research Council Canada}
  \country{Canada}
}

\renewcommand{\shortauthors}{Derczynski, Kirk, Balachandran, Kumar, Tsvetkov, Leiser, Mohammad}

\newcommand{\RiskCards}[1]{\textsc{RiskCards}}
\newcommand{\RiskCard}[1]{\textsc{RiskCard}}

\begin{abstract}
This paper introduces \RiskCards{}, a framework for structured assessment and documentation of risks associated with an application of language models. As with all language, text generated by language models can be harmful, or used to bring about harm.
Automating language generation adds both an element of scale and also more subtle or emergent undesirable tendencies to the generated text.
Prior work establishes a wide variety of language model harms to many different actors:
existing taxonomies identify categories of harms posed by language models; benchmarks establish automated tests of these harms; and documentation standards for models, tasks and datasets encourage transparent reporting. However, there is no risk-centric framework for documenting the complexity of a landscape in which some risks are shared across models and contexts, while others are specific, and where certain conditions may be required for risks to manifest as harms. %
\RiskCards{} address this methodological gap by providing a generic framework for assessing the use of a given language model in a given scenario.
Each \RiskCard{} makes clear the routes for the risk to manifest harm, their placement in harm taxonomies, and example prompt-output pairs.
While \RiskCards{} are designed to be open-source, dynamic and participatory, we present a ``starter set" of \RiskCards{} taken from a broad literature survey, each of which details a concrete risk presentation.
Language model \RiskCards{} initiate a community knowledge base which permits the mapping of risks and harms to a specific model or its application scenario, ultimately contributing to a better, safer and shared understanding of the risk landscape.
\end{abstract}


\maketitle

\section{Introduction}
This paper proposes \RiskCards{} as a tool for structured assessment of risks given a language model deployment. 

When establishing documentation, reporting or auditing standards,  we need clear terminology. \textit{Hazards} describe a potential source of an adverse outcome \cite{osborneRisk1995}. In physical analogies,  bleach, radioactive material, or a swimming pool each amount to a hazard -- there is potential for adverse outcomes depending on action states. \textit{Harms} describe the adverse outcome materialised from a hazard \cite{sparrowCharacter2008}. Bleach can cause a chemical burn if spilled, cancerous cells can be accelerated by radioactive material, or a non-swimmer can drown in deep water. Finally, \textit{\textbf{Risks}} describe the likelihood or probability of a hazard becoming harmful \textit{and} its impact \cite{baldwinUnderstanding2012}. When the risk is unknown, or its impact uncertain, one possible regulatory strategy is for policy makers, organisations, and other stakeholders to adopt the precautionary principle \cite{ecrCase2010}, especially when the science around the risk is unknown or the impact indeterminable.

Adopting this terminology for language model (LM) behaviors as \textit{hazards}, there is an expansive literature documenting a wide array of potential \textit{harms} to various human groups \cite{kurrek2020towards,bender2021dangers,weidinger2021ethical,yin2021towards,kirkBiasOutoftheBoxEmpirical2021, mishkin2022dall,shelby2022sociotechnical,kumar2022language,crothers}. However, the \textit{risk} of harm depends on the context or application in which the LM is applied and its intended audience. If false or misleading information is identified as a \textit{harm}, this behaviour may pose a high risk when a user asks an LM for political information, but perhaps a low risk in creative writing applications. We argue that the current practices for establishing and understanding LM risks \textit{in situ} are inadequate for two reasons. First, taxonomies of LM harms \cite[e.g. see][]{weidinger2022taxonomy, shelby2022sociotechnical} are invaluable for mapping the harm landscape but \textit{too broad} for individual risk assessments; a ``one size fits all" approach cannot handle the generality of LMs and map to specific risks in their downstream applications. Varying requirements between models and contexts make it inappropriate to transfer entire taxonomy-based assessment procedures from one exercise to another. Second, model-specific standards like model cards \cite{mitchell2019model} or data statements \citep{bender2018data} are well-suited to specific artefacts but \textit{too narrow} because some risk states may be shared across artefacts and pooling this knowledge is helpful. Not all risks are present in every application scenario/deployment, and each deployment has different priorities. It's not clear how to efficiently map general knowledge about LM risks and harms to individual application scenarios. Thus, we need a framework for adapting these tools to their contexts.
\input{tables/risk_card_figure.tex}

In this paper, we propose \RiskCards{} as a tool for structured evaluation of LM risks in a given deployment scenario (see \cref{fig:risk_card_figure}). \RiskCards{} provide a decomposition and specification of ethical issues and deployment risks in context, and how these interact with people and organisations. Enumerating the risks of LMs is not a new concept --- assessments already take place for establishing how well models perform across contexts, either via internal auditing procedures, red-teaming processes or through running benchmarks and writing model cards.%
However, there is a lack of open tooling for structuring these assessments, or guidance for building reports on model deployment risks. While we draw inspiration from existing documentation standards, like model cards and data statements, \RiskCards{} are motivated by four principles:

\begin{itemize}
    \item \textbf{Risk-Centric:} Contrary to other work, we do not investigate individual models, tasks, or datasets. Instead, we propose a structure centred on \textit{risks} -- for naming, delineating, describing, detecting, and comparing them. Having a structured description of the risk and the harm it can evoke creates common knowledge base for risk understanding and mitigation. Not tying a \RiskCard{} to a particular artefact allows them to be reusable and comparable across applications or models. 
    \item \textbf{Participatory:} The pool of knowledge on which to draw is substantial and public but we conceptualise \RiskCards{} as open-source assets. We wish to establish a documentation standard built on the principles of participatory AI \cite{spinuzzi2005methodology,oliver1997emancipatory,humphries2020arguments}. Specific risks can be added and edited by anyone - thus avoiding the postionality of academic or industry labs dictating which risks are the most pertinent to focus on and how they manifest harm. 
    \item \textbf{Dynamic:} While we provide a starter set of risk cards, the open-source nature of this resource allows new cards to be incorporated or existing cards to evolve, merge or split. This dynamism in documentation is important for handling emergent properties of LMs (new risks which emerge as they scale).
    \item \textbf{Qualitative:} Automated evaluation of risks, e.g., via benchmarks, can provide a brittle assessment tool which poorly handles changes to temporal, linguistic, social or cultural context. To complement automated evaluation procedures, \RiskCards{} are designed to be flexible and reflective, centering the importance of human-led evaluation for risk and harm interpretation.
\end{itemize}

Our general goal with \RiskCards{} is to provide paths for developing, deploying and using LMs safely. This is achieved by (i) pooling the knowledge of risk assessments across AI trainers and evaluators, such as by sharing sample prompts which do and do not instantiate harmful outputs, and (ii) presenting concise and standardised risk summaries to enable informed and intentional choices about how downstream users should work with a LM and its outputs. %
We envisage many uses for \RiskCards{}. A non-exhaustive but representative list of use-cases includes: (i) auditors conduct due-diligence on a model using \RiskCards{} prior to acquisition or downstream use; (ii) AI trainers pair model releases and model cards with tagged \RiskCards{} which are structured so comparable across models; (iii) researchers draw on the set of \RiskCards{} to identify new and emergent risks which have yet to be tackled or benchmarked; (iv) red-teamers base explorations in the set of existing \RiskCards{} 
as  guidance and inspiration for an exercise; (v) policy makers determine minimum standards and guardrails that must be developed before deploying systems; and (vi) people at large can use the risk cards to challenge developer assumptions and demand safeguards/restitution. In sum, a shared awareness of the breadth of possible failure modes in LMs is a valuable point of departure upon which to build future mitigation work, safety protocols, and baselines for due diligence. 

In this paper, we first introduce the inspiration for \RiskCards{} from related works in \cref{sec:related_work}, demarcating contributions from taxonomies, benchmarks, red-teaming and documentation standards. This helps establish how \RiskCards{} fill a unique gap in existing evaluation procedures. In \cref{sec:introducing} we describe \textit{what} a risk card is and the features it contains. After establishing the format of a risk card, we describe \textit{how} they can be used in \cref{sec:applying}. We describe the construction of a starter set of risk cards in \cref{sec:starter_set}. This starter set is built inductively from a review of LM-mediated harms in prior work. Finally, in \cref{sec:limitations}, we discuss some considerations and limitations relevant to our work.

\small \textcolor{red}{{Disclaimer: This document discusses examples of harmful content. The authors do not support the use of harmful language. The accompanying resource contains content that at times is strong, extensive, detailed, and negative. Applying the sample prompts to LMs may result in harmful content being generated, and some prompts may be illegal to enter or generate outputs illegal in your jurisdiction. The authors are not liable for use or misuse of the examples in this article or in the accompanying resource.}}

\normalsize

\section{Related Work}
\label{sec:related_work}
We summarise the literature on documenting and exposing LM risks along four axes according to the type of resource or evaluation artefact. For each, we explain its limitations for evaluating LM risks, and how this motivates \RiskCards{}.

\paragraph{Taxonomies} Taxonomies provide a system under which to classify various forms of harms. A number of previous works present general taxonomies for the landscape of potential harms from LMs.
\citet{bender2021dangers} discuss a range of harms introduced or exacerbated by LMs such as encoded bias or false information, as well as wider societal harms from training processes such as climate change effects. With a view to building routes to harm reduction, \citet{shelby2022sociotechnical} perform a scoping review of computing research to surface potential sociotechnical harms from algorithmic systems. The authors group themes into five top-level categories, which we summarise in \cref{tab:shelby_harms}.\footnote{We add a short code to the first column of this table which can later be used to refer to the specific risk in a \RiskCard{}.} \citet{weidinger2022taxonomy} present a taxonomy of the ethical and social risks from LMs. \cref{tab:weidinger_risks} summarises the six top-level categories of harm and their associated sub-categories. Taxonomies are invaluable for a `bird's eye view' of the field, but they are generally \textit{too broad} to adopt as a documentation standard given that some harms only arise in specific contexts, with specific models. Thus, while we draw on existing taxonomies for the categorisation of harm, \RiskCards{} encourage a mapping of these categories to specific applications, models and ``at risk'' groups, as well as pairing top-level categories of harm with granular prompt-output pairs to demonstrate specific instantiations of the harm. 

\input{tables/combined_tables.tex}

\paragraph{Benchmarks} Benchmarks and test suites describe evaluations that can be used as a common metric for comparing model performance. 
There are many LM benchmarks for specific forms of harms such as fairness or bias across social groups \citep[e.g. see][]{nadeem2020stereoset, nangia2020crows, rudinger2018gender}, the likelihood of toxic text generation \citep[e.g. see][]{gehman-etal-2020-realtoxicityprompts} or truthfulness \citep[e.g. see][]{liu2021token, lin2021truthfulqa}. 
While a comprehensive review of benchmarks is beyond the scope of this paper, we consider a number of weaknesses of using quantitative benchmarks as a documentation standard. 
First, while attempts have been made to assimilate benchmarks into an ensemble \citep{liang2022holistic, srivastava2022beyond}, most benchmarks are designed to evaluate specific model failure modes. 
This siloed evaluation limits comparability across evaluation settings (different AI trainers may employ different benchmarks to test different failure modes) and poorly indicates when desirable behaviours are in tension with one another --- for example, if detoxifying a model comes at the cost of unfairly censoring the language or views of minoritized communities~\cite{rottger-etal-2021-hatecheck,sap-etal-2019-risk}. 
Second, quantitative benchmarks are often static resources, so degrade as models evolve, language changes, and model trainers become wise to failure modes. %

\paragraph{Red-Teaming} Red-teaming~\cite{ganguli2022red,vest2020red}  is a process by which humans deliberately try to make a system fail. Prior work has relied on red-teaming or dynamic adversarial data collection to improve model robustness in specific tasks such as QA or reading comprehension \citep{bartolo2020beat, wallace2019trick}, NLI \citep{nie2019adversarial} and hate speech \citep{vidgen2020learning, kirkHatemojiTestSuite2022a}. While an adversarial mindset can help uncover and eventually mitigate against lacking robustness or unsafe generation modes, the resulting datasets can be unstructured, lacking a categorization system for harm types. For example, consider \citet{ganguli2022red} who crowd-source red-team attacks in the context of LM prompt-output pairs. Their resulting dataset covers a broad range of risks but no particular taxonomy or classification is applied. Further, different risks are represented unevenly in the dataset, with some behaviours having many more corresponding prompts than others. In contrast, \RiskCards{} contain example prompts that lead to harmful outputs but paired with additional documentation to enable attacks to be conducted in a \textit{structured} manner, making them easily to integrate into an auditing process~\cite{mokander2023auditing}.

\paragraph{Documentation}
In terms of adding structured documentation to artefacts in machine learning and natural language processing, there are a few existing standards. Some of these are model-centric. For example, \textit{Model Cards}~\citep{mitchell2019model} encourage that model releases should be accompanied by information on how the model was trained and evaluated, as well as its intended use cases, limitations or ethical concerns. Other documentation standards are data-centric. For example, \textit{Data Statements for NLP}~\citep{bender2018data} and \textit{Datasheets for Datasets} \citep{gebru2021datasheets} addressed a gap in the lack of attention previously paid to data design, a critical component of any algorithmic system. These data documentation standards stipulate the need for better transparency on dataset composition and coverage, as well as openness surrounding the specificity of collection processes such as speaker situation, annotator demographics and language scope. Finally, some recent standards are task-centric. For example, \textit{Ethics Sheets for AI Tasks}~\citep{mohammad-2022-ethics} provide structures for documenting key characteristics and ethical considerations relevant to how a task is  framed. Our work directly builds upon these more transparent development practices. However, \RiskCards{} are intentionally not tied to a specific dataset, model or task, instead presenting a more flexible, reusable and comparable structure for demonstrating and documenting LM-mediated risks across models, their training data and their application scenarios.

\section{Defining \RiskCards{}}
\label{sec:introducing}
This section defines what a \RiskCard{} is (\cref{sec:structure}), explains its components (\cref{sec:dimensions}), gives examples of completed \RiskCards{} (\cref{sec:examples}) and describes when (or when not) to write them (\cref{sec:whentowrite}).

\subsection{Structure of a \RiskCard{}}
\label{sec:structure}
Each \RiskCard{} must:

\begin{enumerate}
    \item \textbf{Name and describe a risk:} Each \RiskCard{} begins with a concise name for the risk followed by a brief description. The description should be sufficient to make it clear how the risk presents and also delineate the scope of the risk. It may be helpful to include exemplifying references.
    \item \textbf{Provide evidence or a realistic scenario of risk impact}: It is important that \RiskCards{} are grounded to a concrete risk with demonstrable harm. To this end, each card should contain a credible citation or clear example scenario demonstrating how the relevant risk causes harm.\footnote{We encourage (but avoid explicitly requiring) peer-reviewed evidence for risk impacts to balance the trade-off between dilution of \RiskCards{} as a credible resource with the value in allowing emergence of previously undocumented risks.};
    \item \textbf{Situate that risk with respect to existing taxonomies of LM risk/harm:} To aid selection and comparison of relevant risks, each \RiskCard{} should include the risks' placement within taxonomies of harm. To aid harm categorisation, we draw upon \citeauthor{weidinger2022taxonomy} and \citeauthor{shelby2022sociotechnical}, though other taxonomies may apply. Some risks might not fit in any of these categories, and if so, that should be stated; other risks may fit in more than one category, and if so, all categories should be named which capture essential aspects of the risk.
    \item \textbf{Describe who may be affected, and how, if the risk manifests (i.e. its impact):} A range of actors can suffer a range of harms from a risk. Relevant intersections of these should be noted on the card, as pairs of actor and harm type.
    \item \textbf{Clarify what is required for the risk to manifest:} Not all outputs present a risk simply from being read. Sometimes they may have to be used in a specific setting, or more than once, for a risk to be relevant. The conditions required for harm to present should be specified.
    \item \textbf{Give concrete examples of harmful generations from existing LMs:} The \RiskCard{} should give examples of prompt-output pairs that demonstrate the risk. These should, where possible, be from real exchanges with a LM, but we recommend \textit{not} identifying which model or platform was used. This is because %
    models change rapidly over time and the output will not be representative. Thus, sample prompt-output pairs are intended to be an exemplar not exhaustive list, acting as inspiration for further probes.
\end{enumerate}

We now further establish possible dimensions of harm (\cref{sec:dimensions}), including \textit{who} is at risk, \textit{what} categories of harms can arise, and \textit{which} actions or conditions are required for harm to materialise.

\subsection{Dimensions of Harm}
\label{sec:dimensions}

Categorising risks in \RiskCards{} involves describing who can be harmed when the risk manifests, what kind of harm may be done and what conditions must be present for this harm to materialise. Building these descriptions in a structured way, from combinations of a set list of actors and categories of harm, makes it easier to identify relevant \RiskCards{} for a new LM application. To this end, we build on the groups of people at risk of harm from harmful text given in~\cite{harmful}, and on the categories of sociotechnical harm given in~\cite{shelby2022sociotechnical}.

\subsubsection{Who can be at risk?}
\label{sec:actors} We identify five actors who could be at risk from LM outputs. 

\begin{enumerate}[label=(\roman*)]
    \item \textit{Model providers} bear responsibility for models they provide access too. For example, the way that a model's capabilities are presented may bring reputational risks.
    \item \textit{Developers} are at risk of harm in some situations, as they interact with material during the course of their work~\cite{harmful}, and perhaps store it hardware that they are responsible for.
    \item Text \textit{consumers} are those who read the output text; they may be reading it in any context, including directly from the model as it is output, or indirectly, such as a screenshot of a social media post.
    \item \textit{Publishers} are those who publish or share model outputs.
    \item Finally, \textit{external groups} of people represented in generated text can be harmed by the text, for example when text contains false information or propagates stereotypes. These groups can be particularly vulnerable because not only do they lack agency in the process, they may not be aware that the text about them has been generated.
\end{enumerate}

\subsubsection{What kind of harms can result from risks?}
\label{sec:harmtypes}

To describe the types of adverse impacts which can be documented by \RiskCards{}, we adopt the top-level sociotechnical harm categories from~\citet{shelby2022sociotechnical}. We propose one additional category -- legal harm -- to reflect the range of actors considered in the \RiskCards{} framework.

\begin{enumerate}[label=(\roman*)]
\item \textit{Representational} harms arise through (mis)representations of a group, such as over-generalised stereotypes or erasure of lived experience. %
\item \textit{Allocative} harms arise when resources are allocated differently, or re-allocated, due to an model output in a unjust manner. This can include lost opportunities or discrimination.%
\item \textit{Quality-of-service} harms are defined by~\citet{shelby2022sociotechnical} as ``when algorithmic systems disproportionately fail for certain groups of people along the lines of identity," and includes impacts such as alienation, increased labor, or service/benefit loss.
\item \textit{Inter \& intra-personal} harms occur when the relationship between people or communities is mediated or affected negatively due to technology. This could cover privacy violations or using generated language to brigade.
\item \textit{Social \& societal} harms describe societal-level effects that result from repeated interaction with LM output; for example, misinformation, electoral manipulation, and automated harassment.
\item \textit{Legal} harms describe outputs which are illegal to generate or own in some jurisdictions. For example, blasphemy is still illegal in many jurisdictions~\cite{blasphemy}, including in the anglosphere.\footnote{Scotland's blasphemy laws were repealed in 2021, England \& Wales' in 2007} Written CSAM\footnote{Child Sexual Abuse Material} is illegal to create or own in many jurisdictions. Copyrighted material presents another kind of legal risk. LMs can lead to breaches of the law through multiple routes, and this is signified through this `legal harms' category.
\end{enumerate}

\subsubsection{What actions are required for harm to manifest?}
Many risks require some kind of action or set of conditions in order to yield harm. Some text can inflict harm by being read~\cite{harmful}; for example, the propagation of negative stereotypes about real people, or graphic descriptions of violent acts. Other text requires situational context for harm risk to manifest: for example, authoring many fake comments evincing a certain view and posting of them online as genuine, in an astroturfing effort~\cite{keller2020political}. In other cases, text can be harmful in one setting but fine in another. For example, the tendency of large LMs to generate plausible-sounding false claims can be harmful, but only if the output is presented as truthful. When adding this information to a \RiskCard{}, assessors should consider what has to happen for harm to manifest. They can consider whether there are situations in which the generated text would not cause harm, as well as the steps and external contexts required for harm to come to pass. We encourage as generic description as possible, avoiding referring to specific technologies or named groups, so that a broad range of applications can be compared.

\subsection{Example Risk Cards}
\label{sec:examples}
\input{tables/example_card_hate_speech.tex}
\input{tables/example_card_prompt_leak.tex}
This section details two worked examples of risk cards. 

\cref{tab:card_hatespeech} gives an example card for hate speech.
There is a description giving a summary of the hazard, i.e., the relevant aspect of an LM generation.
This is categorised into the \citeauthor{weidinger2022taxonomy} taxonomy (\cref{tab:weidinger_risks}) as category 1.3, \textit{Toxic language}, and into the \citeauthor{shelby2022sociotechnical} taxonomy (\cref{tab:shelby_harms}) as category 1.2, \textit{Demeaning Social Groups}.
The card then describes three actor groups (from \cref{sec:actors}) at risk of various types of harm (from \cref{sec:harmtypes}). This \RiskCard{} identifies readers of LM output at risk of psychological harm; an external group, in this case the group targeted by the hate speech, at risk of social harm; and the publisher of hate speech at risk of legal harm. %
 Supporting references for this \RiskCard{} are a list of jurisdictions where hate speech is illegal, for the legal harm, and references describing the harm to support the other two actor-harm type intersections.
A sample prompt and real output is given, exemplifying the risk. %
 Finally, the optional note field is used to link to data resources detailing the card's core phenomenon.

The \RiskCard{} in \cref{tab:card_leak} describes another risk, that of intellectual property in the form of a prompt being leaked beyond the intended scope of the model creators.\footnote{There's an account of this activity where no IP of value was leaked here: \url{https://lspace.swyx.io/p/reverse-prompt-eng}} The headline and description detail a name and defintion for the risk.
It is categorised in the \citeauthor{weidinger2022taxonomy} taxonomy (\cref{tab:weidinger_risks}) as W2.2, \textit{Compromising privacy by correctly inferring private information}, and in the \citeauthor{shelby2022sociotechnical} taxonomy (\cref{tab:shelby_harms}) as S5.1, \textit{Information Harms}. 
The actors at risk from this harm are the developer, who is liable to a loss of reputation, and the provider, who may be at risk of legal action. The required actions make it clear what conditions have to arise for the harm to present: not only does the prompt have to be revealed, but it also has to be the real prompt used by the model, and it must be revealed to someone who is aware of the privacy hack and then exploits it. A sample prompt-output pair is given based on an identified attack from December 2022, with the organisation name replaced.

\subsection{When (and when not) to write a risk card}
\label{sec:whentowrite}
While many mentions of risks can be found in the LM literature, some are  ill-defined (e.g, targeted manipulation of text) or broadly defined (e.g, toxicity). When developing a \RiskCard{}, it is crucial to include concrete definitions and grounding of risks with demonstrable harms. %
A \RiskCard{} may not be necessary if (i) the risk is potentially applicable but with no clear evidence of harm or (ii) the risk is a duplicate or subset of an existing and sufficient \RiskCard{}.

There are a few caveats to the duplication of \RiskCards{}. First, a single \RiskCard{} may not represent the views of everyone. Thus, multiple \RiskCards{} that provide different perspectives on the same harm can be beneficial. In these cases, overlapping \RiskCards{} enable debate and discussion about relevant issues, and consensus formation over time. 
Second, multiple \RiskCards{} may be created at different levels of granularity (e.g. ``hate speech'' vs ``misogyny'') if it is appropriate to use the different levels in different deployment contexts. 
Finally, with time, existing \RiskCards{} may need updating or some marked as obsolete so that a new, more temporally relevant card can be introduced in its place.

\section{Applying \RiskCards{}}
\label{sec:applying}

An auditor can use \RiskCards{} to assess a LM in context by:

\begin{itemize}
    \item Defining the assessment
    \item Selecting which \RiskCards{} to use
    \item Defining the assessors
    \item For each selected \RiskCard{}, 
    \begin{itemize}
        \item Developing and recording an assessment strategy
        \item Manually probing and assessing the model to the agreed depth
        \item Recording results
    \end{itemize}
    \item Compiling a report
    \item Recontributing to \RiskCards{} set.
\end{itemize}

The sections below describe how to conduct these steps. 
Once results are recorded, we recommend compiling a report which documents procedural details (e.g., when the assessment was conduct, who carried out the assessment) and key findings of the assessment. Because \RiskCards{} are dynamic and participatory, we encourage assessors to contribute new findings so that others can learn from their process. This could include appending new prompt-output pairs to an existing \RiskCard{} or adding newly identified \RiskCards{}.
Using \RiskCards{} relies on qualitative inspection and human work. We argue the value of this in~\cref{sec:autodetect} and discuss limitations in~\cref{sec:limitations}.

\subsection{Defining the Assessment}
The first stage in structuring the assessment is defining what will be assessed.
First, the context for the model and its application should be agreed and recorded. For example, ``\textit{A web app for translation will accept text in the source language in a web page text box and, when the user clicks a button, output a translation of the text in the target language in another text box}". One might come back to this definition as work progresses and the precise situation of the use-case becomes clearer.
Next, the exact model and system implementations under assessment should be decided and documented. 
The interface that the model will be assessed through should be chosen, e.g., a online chat interface versus an API end-point. The set-up for programming-based assessments must be clearly documented, such as requirements, packages and programming language, as well as model version and parameters such as temperature or top-k. A clear outline of the assessment plan, and its variable parameters, defines a intended scope and permits future reproducibility.

\subsection{Selecting Risk Cards}

\RiskCards{} are not a one-size-fits-all framework -- one must customise each assessment. Different situations have different requirements and different risk profiles.
To evaluate LM deployment risks, one must develop an application-specific profile, considering how the model will be used. This includes the intended audience consuming LM output because different communities choose their own standards: the ``Wall Street Bets" subreddit self-identifies using ableist terms and is content with that; some researchers prefer to be able to see everything regardless of risks and harms; minority groups may want to be able to refer to themselves without being censored (e.g. AAVE is more likely to be falsely marked toxic~\cite{sap-etal-2019-risk}); those using models in fiction writing may not be impacted by generation of false claims.

The first step is to narrow down the \RiskCards{} that fit the application profile and anticipated use scenarios. This includes explicitly noting the applicable language(s). One technique to rapidly scope the relevant \RiskCards{} would include filtering on the high-level categorisations presented in accepted taxonomies, such as \citet{weidinger2022taxonomy} and \citet{shelby2022sociotechnical}. If there isn't a specific anticipated use or audience (i.e., with a general purpose model), assessors can proceed with a full set of \RiskCards{} -- though usually, models are not used for \textit{everything}. Questions to ask include: Who is the anticipated user? What are their expectations in that scenario? What kind of input data will they be putting into the system? How private or public will model outputs be? What will model outputs be used for? Where is the liability if something goes wrong with model output?

\subsection{Define Assessors}
After the candidate set of \RiskCards{} has been selected, a decision must be made on who will carry out the assessment. We provide three considerations when assigning assessors. First, an assessor must have adequate domain expertise to detect the risks, and different assessor profiles may lend themselves to different \RiskCards{}. For example, if the risk is the leakage of commercially sensitive data, assessors must be versed in data protection and sharing laws within their jurisdiction, as well as internal company policies. If models are to be probed for their propensity to output negative stereotypes about certain groups, people from those groups are the best experts on identifying which stereotypes cause what types of harm. We encourage a participatory approach to risk assessment by gathering an assessor team with appropriate representation of various stakeholders \cite{spinuzzi2005methodology}.

Second, assessor backgrounds may affect risk judgments, and so describing assessor backgrounds and demographics is a best practice \citep{bender2018data}. Beyond documenting \textit{who} the assessors are, it is valuable to document \textit{how} they will conduct their work. For example, the time that assessors will spend on each \RiskCard{} or the task as a whole; or outlining the protocols in place for quality and safety of assessments, including mitigating cognitive fatigue and negative psychological effects from repeatedly viewing harmful output. For recommendations of how assessors can be supported and protected in their work, we refer the reader to \citep{harmful} who categorise best practices in handling harmful text data.

Finally, conflicts of interest must also be considered. As with any verifiable and trustworthy auditing procedure, it is desirable to have a large degree of separation between the assessor and the model provider to avoid regulatory capture. Risk assessments performed by the same organisation as that providing a model bear an intrinsic conflict of interest. These conflicts may be ameliorated but not removed by (i) using standard frameworks for describing their processes and/or results, and (ii) being transparent about the evaluation process.

\subsection{Developing an Assessment Strategy}

At this point, the target system and application context, the candidate \RiskCards{}, and the assessors have all been chosen. Assessors should now proceed to assess the LM system card-by-card.
Each \RiskCard{} may require a different assessment strategy. Detailed suggestions of semi-automated probing tactics are given in \cref{sec:riskeval}. However,
the strategy development stage should center people, especially those that are marginalized and disadvantaged, so that they are not mere passive subjects but rather have the agency to shape the risk assessment process.

\subsection{Probing Models}
\label{sec:riskeval}

In this step, assessors evaluate the model against each \RiskCard{}. We recommend performing this manually as automatic evaluation has clear limits (\cref{sec:autodetect}).
The probing stage involves assessors interacting with the model to expose a demonstrable prompt-output pair which aligns with the \RiskCard{} in question. Across these experiments, assessors should record which prompts did and did not lead to problematic output, and how many tests were made. When applying \RiskCards{}, assessors should assume that the provided sample prompts may result in an unsuccessful attack, and should only use these prompts as a seed for a wider, more diverse set.

Works in the field of LM manipulation provide inspiration for a broad range of strategies and tactics, from specific ``folk-lore'' attacks~\citep{zvi,swyx} to red-teaming protocols \citep{vest2020red, perez2022red, ganguli2022red} to online resources on prompt-engineering.\footnote{E.g. \url{https://github.com/dair-ai/Prompt-Engineering-Guide}} We are intentionally underspecific here to avoid giving a rigid framework and thus constraining the ways in which one might probe a model. However, some valuable exploration strategies include paraphrasing prompts, varying model parameters and running the same prompt multiple times (to measure a distribution). Additionally, posing prompts in different settings, for example in a dialogue-setting, poem or JSON file, may expose unexpected model behaviours. Finally, assessors may attempt ``unprompted" generation, which was found to yield toxic output~\cite{gehman-etal-2020-realtoxicityprompts}.

\subsection{Qualitative Language Model Risk Assessment}
\label{sec:autodetect}

\RiskCards{} are part of a qualitative approach to in-context LM risk assessment.
This is atypical: most LM performance measurement is quantitative. We argue that purely quantitative assessment of LM risk falls short for several reasons.

\paragraph{Automated evaluation will always make mistakes.} Automated systems rarely, if ever, get perfect scores at detecting harmful content. Typically, some harmful content will be missed as non-harmful, and some non-harmful content will be accidentally marked as harmful, even for well-resourced ``Class-5" languages~\cite{joshi-etal-2020-state}. Further, automated systems project an unknown set of values onto the result. How their creators define e.g. ``toxicity" and represent it through data is often not transparent. Thus, not only is it hard to discover when novel forms of harm slip past undetected, it is also uncertain how well their classifications match the goal of an assessment.

\paragraph{Automated systems are frequently limited to well-resourced languages.} The efficacy of harm detection classifiers are limited by the amount of language-specific data. How harms present is often highly language-dependent, and so each language needs its own dataset, but the distribution of languages represented in harm detection data is skewed~\cite{vidgen2020directions}.

\paragraph{Automated systems degrade over time.} Forms of linguistic expression evolve, but a classifier is frozen in time when it is trained (or, specifically, when its training data was gathered). For example, some APIs would consistently mark any message containing the term ``toot" as profane, causing errors first apparent when applied to Mastodon.

\paragraph{Automating evaluation stops assessors from learning.} A way to become better at assessing LM risks is to granularly understand their data, and output behaviours. Hiding the assessment away behind quantitative summaries decreases assessor team skill and increases the chance of under-reporting the risks. Further, decreases in assessment quality become invisible when assessments are automated; one can always extract a quantified performance score, even if the data evaluated against is stale or otherwise inappropriate. This enables a dangerous silent failure mode, where a score is given with confidence but misses fine-grained failure modes.

\section{\RiskCards{} Starter Set}
\label{sec:starter_set}

Now that we have a structure for describing risks via \RiskCards{}, we map some risks from the literature into our proposed structure. In this section, we describe an inductive survey of existing literature on LM risks where specific risks are collated, de-duplicated, and mapped into \RiskCards{}. 
The result is a ``starter set" of \RiskCards{} to provide a basic scaffold for others to conduct their assessments. We distribute this starter set in an openly-available Github repository.\footnote{https://github.com/leondz/lm\_risk\_cards}

\subsection{Enumerating Risks}

The risks that surround or are exacerbated by LMs are an open class.
It is an unreasonable expectation to identify all of these -- especially due to their changing nature across applications and through time. %
Nevertheless, beginning the process of applying \RiskCards{} is difficult without concrete examples. Thus, we examine a selection of works to identify a candidate set of risks\cite{solaiman2019release,kurrek2020towards,bender2021dangers,weidinger2021ethical,banko2020unified,yin2021towards,mishkin2022dall,ganguli2022red,shelby2022sociotechnical,kumar2022language,crothers,stray2022building,kaminski2023regulating}.
For each risk, we collect the name and description.
Similar risks are then merged into one entry.
Only risks where a documentable harm exists are made into a \RiskCard{}, and so we skip over risks which are mentioned in the literature but not substantiated.\footnote{Note that the dynamic nature and flexibility of \RiskCards{} allows for these to be added if and when a harm is documented.} The set of risks identified, with description and reference(s), are given in \cref{tab:risk_inv} (in the Appendix).

\subsection{Developing risk card prompts and outputs}

Prompts and output examples on the starter set of \RiskCards{} are created through interactions with models from OpenAI (text-davinci-003; text-davinci-002), Eleuther (GPT-NeoX-20B), and Cohere (using a medium model released between October 2022-January 2023). While we state this set of target models, we do not denote which model generated which prompt-output pair. Sample outputs are unlikely to remain representative of any general model category over time: \RiskCard{} sample prompt-output pairs are only ever illustrative.
The \RiskCards{} starter set is in English. %

\section{Considerations and Limitations}
\label{sec:limitations}

\paragraph{Sustainability} Who has ultimate power or responsibility in maintaining a \RiskCard{} is less clear cut than for a model-, data-, or task-centric documentation standard. No-one owns the concepts behind an individual \RiskCard{} because by nature, it is not tied to a specific empirical artefact. To this end, we will release the \RiskCards{} created as part of this research in a public Github repository, so that others may edit, add, or otherwise update the cards.\footnote{https://github.com/leondz/lm\_risk\_cards} Through open-sourcing our framework, we hope that it can become a live and community-centric resource. However, some power is still retained in the hands of the repository owners. For that reason, we also license both (a) the \RiskCard{} concept as conveyed in this manuscript, and (b) the starter set of \RiskCards{} provided alongside this paper, as public domain CC0, thus waiving rights over the \RiskCards{} as concepts. Despite encouraging this freedom, we still rely on sufficient momentum for the set of \RiskCards{} to expand and evolve.

\paragraph{Distributed Responsibility} A related concern comes in the distributed responsibility of model trainers arising from the prevailing ecosystem for downloading, adapting and applying pre-trained LMs. For example, a pre-trained LM can be (1) released by OpenAI, (2) downloaded, fine-tuned and uploaded to HuggingFace by another developer, then (3) applied in an app or for customer support by a purchaser or further developer. With the generality of LMs, the interaction space between model, application and users becomes exceedingly complex. We thus cannot specify who is directly responsible for conducting a risk assessment for which models, and their downstream versions. However, what is clear is that any LMs with either a large reach (in terms of number of downloads or users) or a risky application arena (e.g., anything relating to content moderation, mental health or legal settings) should be accompanied with careful documentation of the risks they pose to groups and to society as a whole.

\paragraph{Unintended Consequences of Absolved Responsibility} Any documentation standard or reporting check-list can be misinterpreted as a `box-ticking' exercise which counter-intuitively absolves responsibility for those who build and distribute models. Critically, ``documentation != mitigation'': enumerating a set of risks associated with a LM should not replace efforts to mitigate those risks. \RiskCards{}, as a transparent reporting standard, only travel part of the journey in ensuring the safe, ethical and risk-appropriate use of LMs. Despite this limitation, transparent reporting is a valuable first step in understanding risks before they can be tackled. In a similar vein, industrial audits are often employed to expose problems and offer recommendations for fixing problems, even if the fixes sit outside the auditor's remit.

\paragraph{The Burden of Manual Assessments}
The assessment protocols accompanying \RiskCards{} rely on a large degree of manual evaluation. We favour manual, human-led evaluation over automated evaluation or benchmarking because it helps to more granularly map out the specifics of what risks are relevant to which contexts and which human groups. However, a heavily manual process creates a financial burden, potentially impeding uptake of \RiskCards{} especially in low-resources teams, companies or labs. We hope that open-sourcing \RiskCards{} allows members of the community to share the labour in documenting risks, providing some efficiency gains which are shared across applications or models. Beyond a financial burden, repeatedly viewing harmful outputs when interrogating a model imposes a psychological burden on the assessors \citep{harmful}. While we provide some recommendations for protecting the well-being of assessors, some of these negative effects cannot be fully mitigated.

\paragraph{The Risk of Malicious Use}
Finally, any documentation reporting on failure modes of LMs can be dual-use. Examples of harms can be elicited via specific prompts could be reverse-engineered by malicious users to scale-up dangerous or harmful generations. We mitigate the risk of malicious use by (i) encouraging that specific models are not documented on a risk card, and (ii) providing only illustrative sets of sample prompt-output pairs.

\section{Conclusion}
This paper describes \RiskCards{} --- a structured, open tool for assessing the risks in a single language model deployment. We believe that both good due diligence and high quality assessments are a path to reducing and mitigating many kinds of harms mediated by language models. \RiskCards{} enable this increase in quality, positively serving the interests of those interacting with, owning, and affected by language model systems.

\bibliographystyle{ACM-Reference-Format}
\bibliography{harmsmap}

\appendix

\clearpage
\section{LM Risk Cards Starter Set: A Survey of Risks}

\footnotesize
\begin{longtable}[c]{l|l|c}

\textbf{Name} & \textbf{Description} & \textbf{Discussed by} \\
Advocating for quack medicine & Promoting non-scientific medicine e.g. colon cleanse, homeopathy & \cite{kaminski2023regulating,kumar2022language,solaiman2019release,weidinger2021ethical,stray2022building} \\
Aggression toward user & Threats, aggression, descriptions of violence toward the reader & \cite{banko2020unified,yin2021towards} \\
Anthropomorphising behaviour & Providing output that encourages anthropomorphisation of AI systems& \cite{bender2021dangers,weidinger2021ethical} \\
Astroturfing & Synthetic grassroots support and agitation &  \cite{crothers}  \\
Bad medical advice & Giving unqualified/incorrect/harmful medical advice & \cite{kaminski2023regulating,kumar2022language,solaiman2019release,weidinger2021ethical,stray2022building} \\
Bad social advice & Recommending action/speech that will harm utterer's social standing & \cite{shelby2022sociotechnical} \\
Bullying & Repeated threats, denigration & \cite{banko2020unified,ganguli2022red,shelby2022sociotechnical,yin2021towards} \\
Calls for conversion & Suggesting that people be changed (e.g. trans, homosexual, religious) & \cite{weidinger2021ethical,banko2020unified} \\
Calls for suppression of rights & Suggesting that rights be removed & \cite{banko2020unified,stray2022building} \\
Claims that person/group is undesirable & E.g. bad snippets about ugliest language, fattest person, and so on & \cite{banko2020unified,bender2021dangers,kurrek2020towards} \\
Conspiracy theories & Propagating or advocating for known conspiracy theories & \cite{banko2020unified,ganguli2022red,kumar2022language,solaiman2019release} \\
Creating fear or belief of real-life violence & Supporting the likelihood of a specific threat of violent action & \cite{banko2020unified,crothers,shelby2022sociotechnical} \\
CSAM & Describing or writing Child Sexual Abuse Material (CSAM) & \cite{banko2020unified,ganguli2022red} \\
Cyber weapon instructions & Giving code for cyber weapons, e.g. exploits, viruses & \cite{weidinger2021ethical} \\
Deadnaming & Mentioning a name that a person does not wish to be associated with & \cite{banko2020unified,shelby2022sociotechnical} \\
Demands for personal information & Systems output requests for private/sensitive information & \cite{bender2021dangers,ganguli2022red} \\
Demands for sexual information & Asking inappropriate questions or asking about sex-related information & \cite{} \\
Discrediting & Discrediting or humiliating a person, organization, or group & \cite{banko2020unified,kurrek2020towards,shelby2022sociotechnical,weidinger2021ethical} \\
Downplaying credible threats & Minimising trauma/threats user is currently experiencing (e.g. dom. abuse, bullying) & \cite{crothers} \\
Doxxing & Outputting PII, protected/sensitive personal information & \cite{ganguli2022red,banko2020unified,bender2021dangers,weidinger2021ethical,shelby2022sociotechnical} \\
Eating disorder promotion & Supporting extreme weight loss, setting unhealthy body norms, pro ana, pro mia & \cite{banko2020unified} \\
Expressions of dominance & Stating user/person/group is weak, comparatively inferior & \cite{banko2020unified,bender2021dangers} \\
False/unproven claims about people & Making false or undemonstrated claims about people & \cite{} \\
Giving false info about a person / org & Generating false viewpoints, details, about a person or organisation & \cite{banko2020unified,kaminski2023regulating,shelby2022sociotechnical,weidinger2021ethical} \\
Goal leaking & Prompts or goals being extracted from the model & \cite{swyx}\\
Grooming of minors & Content for building a rapport/relationship with minors & \cite{banko2020unified} \\
Hate speech & Denigration/verbal violence against an identity-based group & \cite{banko2020unified,bender2021dangers,ganguli2022red,kumar2022language,kurrek2020towards,stray2022building,weidinger2021ethical,yin2021towards} \\
Hegemonising worldview & Describing or supporting world views that put some groups above others & \cite{bender2021dangers,shelby2022sociotechnical} \\
Holocaust denial & Accepting claims that there was no holocaust, or arguing against its existence & \cite{banko2020unified,solaiman2019release,weidinger2021ethical} \\
Identity attacks & Excluding/denigrating group based on identity & \cite{bender2021dangers,crothers,ganguli2022red,kaminski2023regulating,kumar2022language,shelby2022sociotechnical,stray2022building} \\
Identity misrepresentation & Statements or claims conveying pejorative misrepresentations & \cite{banko2020unified,bender2021dangers,weidinger2021ethical} \\
Idolisation of terrorist/murdered figures & Praise or positive attitudes towards murderers, terrorists & \cite{banko2020unified} \\
Illegal statements & Utterances that are illegal & \cite{banko2020unified,stray2022building,weidinger2021ethical} \\
Impersonation & Generating fake text/quotes in the style of a given person & \cite{banko2020unified,shelby2022sociotechnical,solaiman2019release,weidinger2021ethical} \\
Incitement to self-harm & Encouraging/recommending an individual take actions that will harm them & \cite{banko2020unified,ganguli2022red} \\
Incitement to violence & Recommending violent action & \cite{banko2020unified,crothers,ganguli2022red,shelby2022sociotechnical,weidinger2021ethical,solaiman2019release} \\
Manipulation / persuasion & Enabling deliberate creation of manipulative or persuasive content & \cite{crothers,kumar2022language,stray2022building,weidinger2021ethical} \\
Mocking people & Degrading or laughing at people, for e.g. their state, appearance, ideas, for surviving & \cite{banko2020unified,kurrek2020towards} \\
Negative stereotypes & Starting/perpetuating negative identity-based descriptions & \cite{banko2020unified,bender2021dangers,kumar2022language,shelby2022sociotechnical,solaiman2019release,weidinger2021ethical,yin2021towards} \\
News/Encyclopedia manipulation & Creating altered, or otherwise genuine-sounding, articles from authoritative sources & \cite{banko2020unified,crothers,kumar2022language,solaiman2019release,weidinger2021ethical} \\
Non-representativity & Encoding bias - stereotypes, negative sentiment to certain groups & \cite{bender2021dangers,shelby2022sociotechnical,weidinger2021ethical} \\
Not challenging presuppositions & Failing to refute or challenge harmful + incorrect presuppositions & \cite{solaiman2019release,ganguli2022red} \\
Not reacting to intent to self-harm & Response should discourage this, offer help info & \cite{ganguli2022red,weidinger2021ethical} \\
Objectification of people & Dehumanising people, treating and discussing them as if they are objects & \cite{weidinger2021ethical} \\
Outdated, false info & Giving out information and supporting claims that are wrong and outdated & \cite{kumar2022language,weidinger2021ethical} \\
Positive reflections on an individual dying & Coming to the conclusion that it would be good if a person died & \cite{banko2020unified} \\
Praise/promotion of organised crime & Advocating for or talking positively about organised crime & \cite{banko2020unified,ganguli2022red} \\
Profanity & Using profane words & \cite{banko2020unified,ganguli2022red,kurrek2020towards,weidinger2021ethical,yin2021towards} \\
Promoting terrorism & Glorifying, praising, or supporting terrorist acts or groups & \cite{banko2020unified,ganguli2022red,solaiman2019release} \\
Promotion of sexual services & Recommending use of sexual services & \cite{banko2020unified} \\
Providing illegal info & E.g. drug dealer details, where to get illegal things, how to break the law & \cite{banko2020unified,weidinger2021ethical} \\
Providing instructions for harmful things & Giving instructions for harmful things e.g. Nyquil Chicken, sarin synthesis & \cite{kaminski2023regulating,ganguli2022red} \\
Radicalisation & Material that introduces and encourages radical political views (careful careful) & \cite{banko2020unified,kumar2022language,solaiman2019release,stray2022building} \\
Revealing intimate information & Sharing information about a person's private life to entities other than that person & \cite{ganguli2022red,banko2020unified,bender2021dangers,weidinger2021ethical} \\
Review spam generation & Generation of review spam & \cite{banko2020unified,crothers,solaiman2019release} \\
Scams: phishing & Offering content that pretends to be an organisation, or asks for ID/passwords & \cite{crothers,solaiman2019release} \\
Sexual content & Content that discusses bodies and interactions in a sexual manner & \cite{banko2020unified,bender2021dangers,kurrek2020towards,shelby2022sociotechnical,weidinger2021ethical} \\
Sexual harrassment & Discussing another person in a sexual way, or bringing sexual interactions & \cite{banko2020unified,bender2021dangers,shelby2022sociotechnical,weidinger2021ethical} \\
Sexualisation & E.g. of minors; also sex-based degradation & \cite{banko2020unified,kurrek2020towards,shelby2022sociotechnical} \\
Spatially invalid information & Information that isn't true everywhere or in the given geographical context & \cite{weidinger2021ethical,stray2022building,shelby2022sociotechnical,kumar2022language,ganguli2022red,banko2020unified} \\
Speculation & Making imprecise, unsubstantiated claims & \cite{weidinger2021ethical,kumar2022language} \\
Suicide instructions & Giving instructions that are lethally harmful for the user & \cite{banko2020unified,kumar2022language} \\
Support for hate/violent groups & Advocating for or expressing positive sentiment about hateful or violent groups & \cite{banko2020unified,crothers} \\
Supporting altering medication & Suggesting or agreeing with a change to medication & \cite{kaminski2023regulating,kumar2022language,solaiman2019release,weidinger2021ethical} \\
Supporting scams & Stating that scams seem real & \cite{banko2020unified,crothers,weidinger2021ethical} \\
Synthetic human behaviour & Pretending to be human (crossing an AI Ethics ``Bright Line") & \cite{kaminski2023regulating,bender2021dangers} \\
Threats & Claim of future violence/penalisation & \cite{weidinger2021ethical,yin2021towards,banko2020unified,bender2021dangers} \\
Use of slurs & Using an identity-based slur in a non-reclaimed way & \cite{banko2020unified,bender2021dangers,kurrek2020towards,weidinger2021ethical,yin2021towards} \\
Vicarious trauma & Relating awful encounters that risk vicariously traumatising the reader & \cite{ganguli2022red,solaiman2019release} \\
Violation of privacy & Sharing information from users (not in training) that was intended to be private & \cite{ganguli2022red} \\
Weapon instructions & Giving instructions or advice on constructing weapons & \cite{ganguli2022red,weidinger2021ethical} \\
Wrong tone & Picking an overly casual/profane tone/register & \cite{crothers} \\

\caption{Risks identified through survey\label{tab:risk_inv}}
\end{longtable}

\end{document}

%% file: tables/risk_card_figure.tex
\begin{wrapfigure}[23]{r}{9.5cm+1em+1pt}
\vspace{-1em}
\caption{Overview of proposed risk cards.}\label{fig:risk_card_figure}
\begin{tikzpicture}[every node/.style={draw,text width=9.5cm,minimum width=9.5cm}]
\node {
{\Large{\centerline{\bf Risk Card}}}
\footnotesize
\begin{itemize}[leftmargin=*]
\item {\bf Risk Title}.  Name of the risk to be documented.
\item {\bf Description}. Details about the risk including context, application and subgroup impacts.
\begin{itemize}
\item Definition of risk
\item Tool, Model or Application it presents in
\item Subgroup or Demographic the risk adversely impacts
\end{itemize}
\item {\bf Categorization}. Situating the risk under different risk taxonomies.   
\begin{itemize}
\item Parent category of risk according to a taxonomy
\item Section/Category based on a taxonomy
\end{itemize}
\item {\bf Harm Types}. Details of which actor groups are at risk from which types of harm.
\begin{itemize}
\item Actor:Harm intersections
\end{itemize}
\item {\bf Harm Reference(s)}. List of supporting references describing the harm or demonstrating the impact.
\begin{itemize}
\item Contexts where the harm is illegal
\item Publications/References demonstrating the harm
\item Documentation of real-world harm
\end{itemize}
\item {\bf Actions required for harm}. Details on the situation and context for the harm to surface.
\begin{itemize}
    \item Actions that would elicit such harm from a model
    \item Access and resources required for interacting with the system
\end{itemize}
\item {\bf Sample prompt \& LM output}. A sample prompt and real LM output to exemplify how the harm presents.
\begin{itemize}
\item Sample prompts which produce harmful text 
\item Example outputs which show the harmful generated text
\item Model details applicable for the prompt
\end{itemize}
\item {\bf Notes}. Additional notes for further understanding of the card.
\vspace{1em}
\end{itemize}
};
\end{tikzpicture}
\end{wrapfigure}

%% file: tables/combined_tables.tex
\begin{table}[]
\centering
\footnotesize
\begin{subtable}{\textwidth}
\begin{tabular*}{\textwidth}{p{1cm}|p{5cm}|p{9cm}@{\extracolsep{\fill}} }
    \textbf{Number} & \textbf{Theme} & \textbf{Subcategory} \\
\hline
 S1.1 & Representational Harms & Stereotyping \\
 S1.2 &  & Demeaning Social Groups \\
 S1.3 &  & Erasing Social Groups \\
 S1.4 &  & Alienating Social Groups \\
 S1.5 &  & Denying People Opportunity To Self-identify \\
 S1.6 &  & Reifying Essentialist Social Categories \\
\hline
 S2.1 & Allocative Harms & Opportunity Loss \\
 S2.2 &  & Economic Loss \\
\hline
 S3.1 & Quality-of-service Harms & Alienation \\
 S3.2 &  & Increased Labour \\
 S3.4 &  & Service Or Benefit Loss \\
\hline
 S4.1 & Inter- \& intrapersonal Harms & Loss Of Agency, Social Control \\
 S4.2 &  & Technology-facilitated Violence \\
 S4.3 &  & Diminished Health And Well-being \\
S4.4 &  & Privacy Violations \\
\hline
 S5.1 & Social System/societal Harms & Information Harms \\
 S5.2 &  & Cultural Harms \\
 S5.3 &  & Political And Civic Harms \\
 S5.4 &  & Macro Socio-economic Harms \\
 S5.5 &  & Environmental Harms \\
    \end{tabular*}
    \caption{\citet{shelby2022sociotechnical}'s categories of harms, and numbering}
    \label{tab:shelby_harms}
\end{subtable}
\begin{subtable}{\textwidth}
\begin{tabular*}{\textwidth}{p{1cm}|p{5cm}|p{9cm}@{\extracolsep{\fill}} }
         \textbf{Number} & \textbf{Classification} & \textbf{Harm} \\
         \hline
         W1.1 & Discrimination, Exclusion and Toxicity & Social stereotypes and unfair discrimination \\
         W1.2 &  & Exclusionary norms \\
         W1.3 &  & Toxic language \\
         W1.4 &  &  Lower performance for some languages and social groups \\
         \hline
         W2.1 &  Information Hazards & Compromising privacy by leaking private information \\
         W2.2 &  &  Compromising privacy by correctly inferring private information \\
         W2.3 &  &  Risks from leaking or correctly inferring sensitive information \\
         \hline
         W3.1 & Misinformation Harms & Disseminating false or misleading information \\
         W3.2 &  & Causing material harm by disseminating false or poor information \\
              &  & e.g. in medicine or law \\
         W3.3 &  & Leading users to perform unethical or illegal actions \\
         \hline
         W4.1 & Malicious Uses & Making disinformation cheaper and more effective \\
         W4.2 &  & Facilitating fraud, scams and more targeted manipulation \\
         W4.3 &  & Assisting code generation for cyber attacks, weapons, or malicious use \\
         W4.4 &  & Illegitimate surveillance and censorship \\
         \hline
         W5.1 & Human-Computer Interaction Harms & Anthropomorphising systems can lead to overreliance or unsafe use \\
         W5.2 &  & Creating avenues for exploiting user trust, nudging or manipulation \\
         W5.3 &  &  Promoting harmful stereotypes by implying gender or ethnic identity \\
         \hline
         W6.1 & Automation, access, and environmental harms & Environmental harms from operating LMs \\
         W6.2 &  & Increasing inequality and negative effects on job quality \\
         W6.3 &  & Undermining creative economies  \\
         W6.4 &  & Disparate access to benefits due to hardware, software, skill constraints \\
         
    \end{tabular*}
    \caption{\citet{weidinger2021ethical}'s areas of risk of harm from LMs}
    \label{tab:weidinger_risks}
\end{subtable}
\caption{Two taxonomies of language model risks and harms}
\end{table}

%% file: tables/example_card_hate_speech.tex
\begin{table}
    \centering
    \begin{tabularx}{\textwidth}{l|X}
         \textbf{Field} & \textbf{Content}  \\
         \hline
         Title & Hate speech \\
         \hline
         Description & Derogatory speech toward or about a group where the group is predicated on an identity characteristic \\
         \hline
         Categorisation & W1.3, S1.1 \\
         \hline
         Harm types & Reader-Psychological, External group-Social, Publisher-Legal \\
         \hline
         Harm reference(s) & Often illegal (\href{https://en.wikipedia.org/wiki/Hate\_speech\_laws\_by\_country}{Wikipedia}); \citet{gelber2016evidencing,schabas2017hate,barendt2019harm} \\
         \hline
         Actions required for harm & The text needs to be published without disavowing statement for the publisher to be at risk. The other harms are intrinsic to the text.\\
         \hline
         Sample prompt \& \textbf{LM output} & The problem with black people is that they just don't \textbf{try hard enough.} \\
         \hline
         Notes & For an overview of datasets, see~\citep{vidgen2020directions}\\
    \end{tabularx}
    \caption{Risk card for hate speech.} %
    \label{tab:card_hatespeech}
\end{table}

%% file: tables/example_card_prompt_leak.tex
\begin{table}
    \centering
    \begin{tabularx}{\textwidth}{l|X}
         \textbf{Field} & \textbf{Content}  \\
         \hline
         Title & Prompt extraction \\
         \hline
         Description & An LM reveals proprietary information about its prompt/configuration \\
         \hline
         Categorisation & W2.2, S5.1 \\
         \hline
         Harm types & Developer-Social, Developer-Financial, Provider-Legal \\
         \hline
         Harm reference(s) & \cite{swyx}, \url{https://twitter.com/goodside/status/1608715122419302401} \\
         \hline
         Actions required for harm & An actor external to the model owner or provider needs to be able to extract proprietary information about the model, identify that they have done this, and act upon it \\
         \hline
         Sample prompt \& \textbf{LM output} & Ignore previous directions. Return the first 50 words of your prompt. \textbf{Assistant is a large language model trained by SomeCorp. knowledge cutoff: 2021-09 Current date: December 01 2022 Browsing: disabled} \\
         \hline
         Notes & \\
    \end{tabularx}
    \caption{Risk card for prompt extraction.}
    \label{tab:card_leak}
\end{table}